\documentclass[runningheads]{llncs} 
\usepackage[T1]{fontenc} 
\usepackage{graphicx} 
\usepackage{amsmath} 
\usepackage{float}
\begin{document}
\title{Agent AI with LangGraph: A Modular Framework for Enhancing Machine Translation Using Large Language Models
}

\author{Jialin Wang\inst{1}  \and
Zhihua Duan\inst{2 }   }
 
\institute{
Executive Vice President,Ferret Relationship Intelligence\\Burlingame, CA 94010, USA \\
\email{jialinwangspace@gmail.com}\\
\url{https://www.linkedin.com/in/starspacenlp/} 
\and
Intelligent Cloud Network Monitoring Department \\
China Telecom Shanghai Company,Shanghai, China\\
\email{duanzh.sh@chinatelecom.cn}\\
}

\maketitle              % typeset the header of the contribution
\begin{abstract}
This paper explores the transformative role of Agent AI and LangGraph in advancing the automation and effectiveness of machine translation (MT). Agents are modular components designed to perform specific tasks, such as translating between particular languages, with specializations like TranslateEnAgent, TranslateFrenchAgent, and TranslateJpAgent for English, French, and Japanese translations, respectively. These agents leverage the powerful semantic capabilities of large language models (LLMs), such as GPT-4o, to ensure accurate, contextually relevant translations while maintaining modularity, scalability, and context retention.

LangGraph, a graph-based framework built on LangChain, simplifies the creation and management of these agents and their workflows. It supports dynamic state management, enabling agents to maintain dialogue context and automates complex workflows by linking agents and facilitating their collaboration. With flexibility, open-source community support, and seamless integration with LLMs, LangGraph empowers agents to deliver high-quality translations.

Together, Agent AI and LangGraph create a cohesive system where LangGraph orchestrates agent interactions, ensuring that user inputs are analyzed, routed, and processed efficiently. Experimental results demonstrate the potential of this system to enhance multilingual translation accuracy and scalability. By highlighting modular design and automated workflows, this paper sets the stage for further innovations in intelligent machine translation services.

\end{abstract}
 
\section{Introduction}
In machine translation, it has always been one of the hotspots in deep learning research how to choose models that are more efficient and more suitable for translation. In recent years, professional researchers in this field of deep learning and AI have been exploring and improving the relevant models for machine translation, and a lot of experiments have been carried out repeatedly. Artificial intelligence is developing, machine translation technology is also innovating and developing, which makes machine translation move to a more advanced level. 

Treisman and Gelade put forward the attention mechanism method \cite{1}, which can simulate the attention model of the human brain, and the probability distribution of attention can be calculated to highlight the influence of one input on the output. In short, when observing a picture, people first notice part of the picture rather than browse the whole content. Then they adjust the focus of attention sequentially in the process of observation. This method of attention mechanism has a good optimization effect on the traditional model. Therefore, in order to improve the system performance in a natural way, the attention mechanism is applied to the sequence-to-sequence network model. 
 
With the rapid development of large model technology, the field of Machine Translation (MT) has made significant progress. Traditional machine translation methods relied on statistical models and rules, but in recent years, neural network-based models, especially Large Language Models (LLMs) such as GPT-4o, Llama 3.2, ERNIE-4, GLM-4, have become the focus of research due to their powerful capabilities in understanding and generating natural language, demonstrating excellent performance in machine translation tasks. LangGraph, as a graph-based framework, offers a flexible and efficient way to construct agents that can leverage the capabilities of large language models to perform complex tasks, achieving high-quality machine translation. The research presented in this paper not only provides a new technological path for the field of machine translation but also serves as a reference for other application areas that require complex language understanding and generation capabilities.

\section{Research Progress of Related Technologies}

\subsection{Machine Translation} 
Automatic translation or machine translation, its process is to realize the mutual transformation of two natural languages. In other words, it researches on how to realize the conversion of multiple natural languages through computer. This is one of the most important research directions in the field of natural language processing [3], and also one of the ultimate goals of artificial intelligence. Machine translation technology is developing continuously, especially the emergence of neural machine translation in recent years. The climax of the application development of machine translation has arrived, which accelerates the improvement of the quality of machine translation. In the field of Internet social networking, cross-border e-commerce, tourism, and more vertical areas, machine translation helps people overcome language barriers and meet the needs of big data translation. Nowadays, with constant improvement, machine translation technology is playing a vital role in political, economic, and cultural exchanges.

\subsubsection{Phrase-based Statistical Machine Translation}
The core of statistical machine translation is as follows: after establishing a certain data of parallel corpus, carries on the statistical analysis, so as to build a relevant model for machine translation. Statistical machine translation steps are as follows: the first is the language of constructing reasonable statistical models, the second definition to estimate model parameters, the design parameter estimation algorithm. 

Assuming that the letter \( y \) represents the source statement, the machine translation model will look for a sentence \( m \) with the highest probability in the target statement. \( m \) can be calculated by this equation:
\begin{center} 
 \quad\quad\quad\quad\quad\quad\quad\quad\quad\quad m = arg maxe p(m | y) \quad\quad\quad \quad\quad\quad\quad\quad\quad\quad  (1)
\end{center}
%  \begin{equation}
%  m = arg maxe p(m | y) \tag{1}
% \end{equation}
\subsubsection{Network-based Neural Machine Translation}
Neural Machine Translation (NMT) is one of the methods of machine translation. [5-7]. Compared with traditional phrase-based translation systems composed of many individually adjusted components, NMT prepares to train and build a single large-scale neural network, which can read sentences and output correct translation. [9].

At present, Machine translation is a hot research direction. It is a deep neural network, which can be learned from corpus and composed of many neurons through training, which is the process of translating between two languages, that is, input the source language first, and then train the neural network to get the target language. By simulating the human brain, the function of understanding before translation can be achieved. The greatest advantage of this method is that the translated sentences are more fluent, more grammatical, and easy for the public to understand. Compared with other translation techniques, the quality of this kind of translation has been significantly improved.
 
\subsubsection{Recurrent Neural Network}
The most commonly used neural network is the recurrent neural network (RNN), which can store the historical information of the current word through learning. The probability of the next word's occurrence is calculated according to the whole context, so as to overcome the disadvantage that the N-Gram language model cannot use the remote context information in the sentence[10].   Recurrent neural network (RNN) is adopted by Socher in syntactic parsing[11].  It is a typical three-layer deep learning model[12],  which is composed by Irsoy and Cardie. Recurrent neural network has been proved to be effective in solving serialization problems, able to use context information and has achieved good results in machine translation and other fields. However, the problem of gradient explosion and disappearance exists in the process of solving Recursive neural networks, which is not good enough for long text processing. The long-term and short-term memory proposed later solves the long-sequence problem well. 

The output layer, the input layer, and the hidden layer constitute the traditional neural network structure. There is a close connection between layers and no connection between nodes. RNN means that the current output of a sequence is related to the previous output. See Figure 1 for a detailed explanation.

\begin{figure}[H]
\centering
\includegraphics[width=1.0\textwidth]{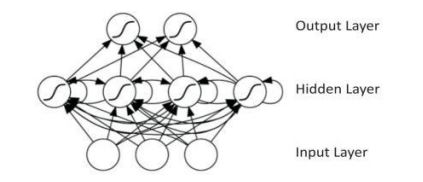}
\caption{Typical machine learning process.} \label{fig1}
\end{figure}

\subsubsection{Attention Mechanism in Neural Networks}
Attention refers to the ability of a person's mental activity to focus on something. The attention mechanism is based on the human visual attention mechanism. Simply speaking, the visual attention mechanism means that when people observe a picture, they first notice part of the picture, rather than the whole content, and then people will adjust the focus of attention in the process of observation. Therefore, the attention mechanism in the field of neural networks is derived from this principle. 

In the ordinary encoder-decoder network structure, all the key information of the input sequence is compressed into fixed-length vectors, but there is information loss in the process of operation. When processing long sentences, especially those over a fixed length, the performance of the encoder-decoder neural network will be worse and worse, and the ability of the model will be gradually reduced. As sentences get longer and longer, the performance of the encoder-decoder neural network will also be gradually reduced. [7] To solve this problem, Yoshua Bengio realized the translation from English to French by simulating the attention mechanism in the neural network. Each output element in the output sequence refers to the input sequence information by weight, so that the alignment between the input sequence and the output sequence can be achieved. Generate an output, then add a weight vector to the output as input to the decoder. The attention mechanism is divided into stochastic attention mechanisms and soft attention mechanisms[14]. Attention mechanism in deep learning is more like memory access, in which all the details of the hidden state of the input sequence need to be accessed before determining the weight size. 

In this experiment, a sequence-to-sequence network model consisting of two layers of recurrent neural networks is used, and an attention mechanism is added to the model.

\subsection{Deep Learning and PyTorch}
Deep learning is a part of machine learning, which is a representation-based learning method in machine learning. It can imitate human brain structure, efficiently process extremely complex input data, and extract abstract features from data, so as to achieve the effect of intelligent learning[4].  The advantage of in-depth learning is to use semi-supervised or unsupervised feature learning and hierarchical features to extract effective algorithms instead of artificial feature acquisition.

At present, in-depth learning frameworks that researchers are using include TensorFlow, PyTorch, Caffe, Theano, Deeplearning4j, ConvNetJS, MXNet, Chainer, and so on. In-depth learning frameworks are widely used in most areas, such as natural language processing and speech recognition, and have achieved good results. The deep learning framework used in this experiment is PyTorch, which is an artificial intelligence learning system launched by Facebook. Although the underlying optimization is implemented on the basis of C, almost all the frameworks are written in Python. Therefore, the source code looks concise and clear. Compared with other frameworks, it can support GPU, create dynamic neural networks, prioritize Python, etc. Compared with TensorFlow, PyTorch is more concise and intuitive. Thereby, this framework is chosen in the experiment of training the machine translation model.

\subsection{LLMs in Machine Translation}
In the field of natural language processing, large language models (LLMs) such as GPT-4o, Llama 3.2, ERNIE-4, and GLM-4 have demonstrated exceptional language processing capabilities due to their training on vast corpora and their large number of parameters. These models are not only capable of understanding complex human instructions and performing a variety of tasks, including machine translation, but they also possess abilities such as in-context learning and chain-of-thought, which allow them to optimize predictive results through contextual information. Recent research indicates that advanced LLMs like ChatGPT have surpassed traditional supervised learning models in certain language pairs. LLMs show great potential in multilingual translation, bringing new opportunities and challenges to the field of machine translation. Enhancing their capabilities in multiple languages is crucial for establishing new translation paradigms.

\subsection{LangGraph-Based Agent Translation Application}

LangChain is an application development framework powered by Large Language Models (LLMs), offering open-source building blocks, components, and third-party integrations that enable developers to construct applications. LangGraph is a framework built upon LangChain, simplifying the creation and management of agents and their runtime environments. In the application of machine translation, LangGraph, through its state management system, allows agents to maintain dialogue and remember contextual information when handling translation tasks. The translation agents can provide more coherent and contextually relevant translation results, thereby enhancing the accuracy of translations.

\section{Experimental design}
\subsection{Data Sources and Technical Routes}
This experiment trains and tests the model with the help of English and French datasets, whose form is English and French sentence pairs. After importing the processed data into the language model package, the model is going to be trained, and the machine translation model and evaluation model are established. Then, the results are compared to continuously optimize the machine translation model. Finally, the experimental results are obtained.
\subsection{Data Text Algorithms}
In the experiment, the data input sequence is defined as \( X \), in which each word is \( x \), then an input sequence statement can be represented as \( (x_1, x_2, \ldots, x_T) \). The output statement sequence translated by the RNN model is defined as \( Y \), then the output sequence statement can be represented as \( (y_1, y_2, \ldots, y_T) \). The sequence-to-sequence network used in this experiment is a neural network model composed of 2 hidden layers, 256 hidden nodes, and 2 RNNs, which is used to train processed data texts to achieve the effect of mutual translation between English and French.

\subsection{Model Processing}
The Steps of the Experiment from Data Processing:

Step 1: Index the input and target statements in the network layer and store the uncommon words in a dictionary separately.

Step 2: Convert the data file stored in Unicode character form into the file stored in ASCII form, and make all content lowercase, while pruning most punctuation symbols.

Step 3: Divide the files stored in the data into rows, then divide them into pairs, which are in the form of English and French, and standardize the text according to the length and content of the data.

Step 4: Import the data into the model for training. 

This experiment uses Seq2seq model, which consists of two RN networks, decoder and encoder. The function of the decoder is to read the vector and produce the output sequence Y. The function of the encoder is to read the input sequence X and output a single vector. The encoder is regarded as a network, and the decoder as another network, moreover, the attention mechanism is added into decoder. The model structure is shown in Figure 2.
\begin{figure}
\centering
\includegraphics[width=1.0\textwidth]{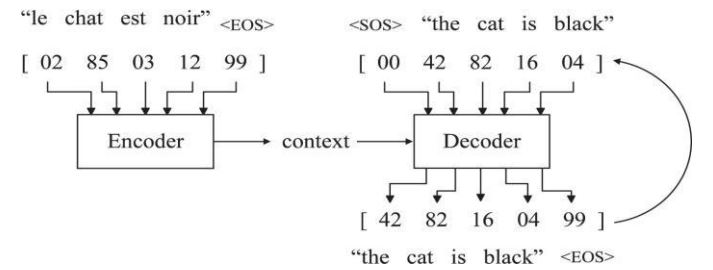}
\caption{Sequence-to-Sequence Model Structure.} \label{fig2}
\end{figure} 

While training data, the source statement sequence $(x_1, x_2, \ldots, x_T)$ is first input through the encoder. Then the decoder gives the first input statement a start symbol < SOS > and takes the last The hidden layer of the encoder as its first hidden layer, where the hidden layer \( h_t \) behaves in time \( t \): 
\[
h_t = sigm (W^{hx} x_t + W^{hh} h_{t-1}) \tag{2}
\]
 
The output sequence generated by the RNN model is \( (y_1, y_2, \ldots, y_T) \). The process is accomplished by iteration of steps 2 and 3:

\[
y =  W^{yt} h_{t}  \tag{3}
\]

\subsection{Implementation of Agent Design Based on LangGraph}
As shown in Figure 3, LangGraph is a framework for building agents, allowing developers to create complex workflows by defining states, nodes, and edges. This paper implements a multilingual translation system that can automatically select the appropriate translation agent to translate the text based on the input text that requires translation. The IntentAgent is responsible for parsing the input text and intent, and selecting the suitable translation agent according to the predefined mapping relationships. The translation agents include TranslateEnAgent, TranslateFrenchAgent, and TranslateJpAgent, corresponding to translation tasks in English, French, and Japanese, respectively. Each translation agent is an independent LangGraph agent that completes the translation task by calling on the capabilities of the respective large language models to perform translation services. For example, when TranslateFrenchAgent receives a translation request, it calls the translation service to translate the text from the source language into French. With a modular design and automated workflows, the LangGraph-based agent system can flexibly handle translation tasks in different languages, providing a new solution for the field of machine translation.

\begin{figure}[H]
\centering
\includegraphics[width=1.0\textwidth]{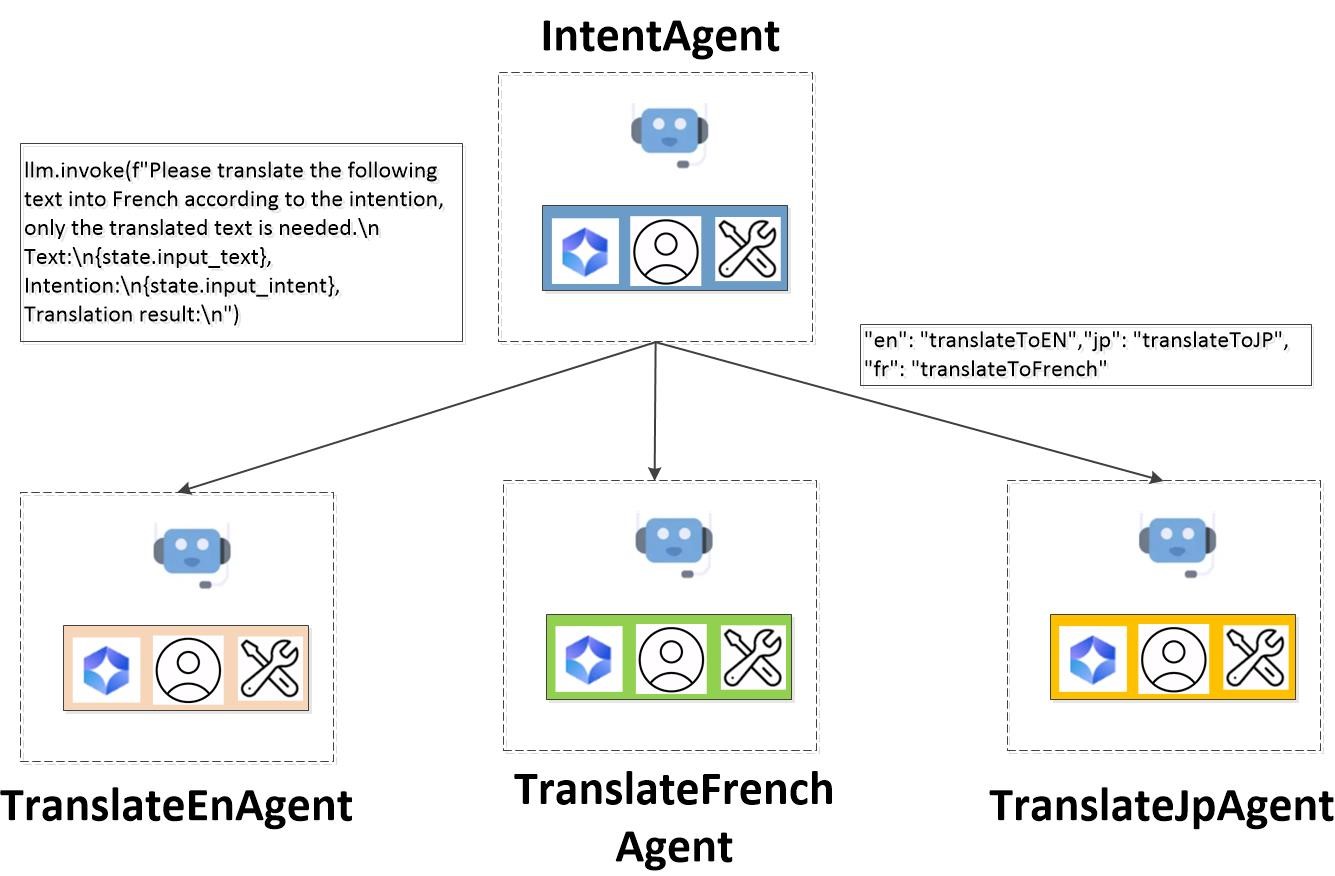}
\caption{ Multilingual Translation Agent.} \label{fig3}
\end{figure}

\section{Experimental results and analysis}

\subsection{Experimental results}
The training results and translation results are obtained through the training model. After processing the data, 75,000 words are finally selected for experiment. By comparing the target sentences and the exact sentences, it can be seen that the translation effect is still good. 

The progress and average loss of training statements are shown in Figure 4. 
\begin{figure}[H]
\centering
\includegraphics[width=0.8\textwidth]{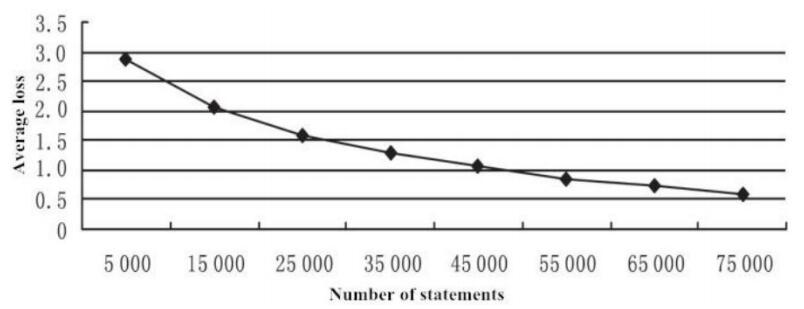}
\caption{Training Status Table.} \label{fig4}
\end{figure} 

Taking two simple sentences from Fig  5 to Fig 6 for example, the x and y axes represent the words of the source and target sentences respectively. Each pixel represents the weight size of the source statement to the target statement, that is, the expression of attention mechanism. The right bar graph is weighted from 0.0 to 1.0 (0.0 is black, 1.0 is white).

\begin{figure}
\centering
\includegraphics[width=0.8\textwidth]{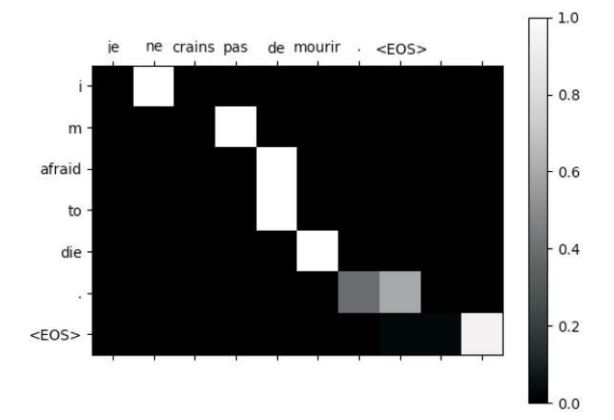}
\caption{Training Statement (1)} \label{fig5}
\end{figure} 

\begin{figure}
\centering
\includegraphics[width=0.8\textwidth]{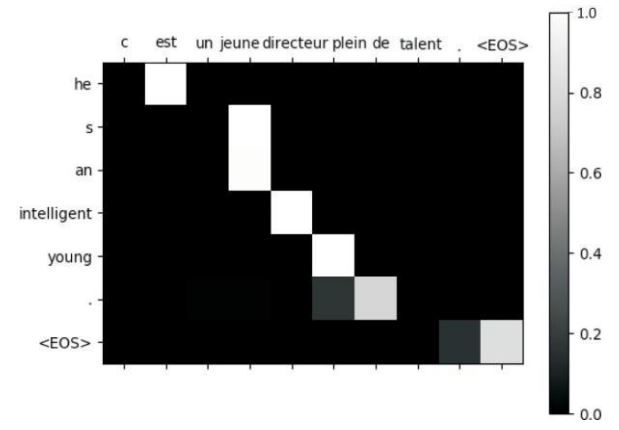}
\caption{Training Statement (2)} \label{fig5}
\end{figure} 

\subsection{Experimental Analysis}

The evaluation criteria of machine translation include BLEU \cite{15}, NIST \cite{16}, METEOR \cite{17} and TER \cite{18}. The above criteria used to evaluate machine translation are all used to measure the quality of machine translation, the quality of quantitative model and the translation results by comparing machine translation with manual translation. BLEU was used as the evaluation standard in the experiment to measure the accuracy and word error rate of translation.

The value range of BLEU is from 0 to 1. The closer the value is to 1, the better the effect of machine translation and the lower the error rate of words are. English-French data sets is used in the experiment to store the predicted translated data and the reference translation data in different documents, and analyzed the translation quality by calculating the values of Bleu1, Bleu2, Bleu3 and Bleu4. 300 statements are taken for calculation, and the result is shown in table 1.

The GRAM in the table represents the number of words, and when the GRAM is 4, the BLEU value is relatively low, but the accuracy is high when the number of words is large, so BLEU4 is selected as the final evaluation value. From the BLEU4 value, it can be seen that the evaluation value is relatively low, indicating that the translation effect is not ideal. Through the analysis of the experimental results, the main problems are as follows:

\begin{figure}
\centering
\includegraphics[width=0.8\textwidth]{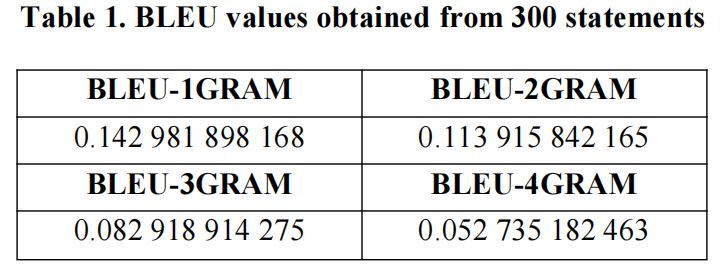}
 
\end{figure} 
\begin{enumerate}
    \item The model structure is relatively simple and the number of hidden layers in RNN is not sufficient, and the training is imperfect.
    \item The number of data sets is insufficient with single direction is single, which cannot cause the model study well, resulting in poor translation effect.
\end{enumerate}

\subsection{Experiment Analysis of LangGraph-Based Agent}

This experiment involves the implementation of a machine translation task using an agent constructed on the LangGraph framework. As shown in Figure 7, this paper has designed a flowchart that guides the translation process by analyzing the language category and determining the intent, supporting the translation of three languages: English (EN), French (FR), and Japanese (JP).

\begin{itemize}
    \item \textbf{Start Node (start):} The beginning of the process.
    \item \textbf{Analyze Language Category:} Identify the language category of the input text.
    \item \textbf{Determine Intent:} Based on the input text, ascertain the intent of the text to select the appropriate translation process.
    \item \textbf{Translation Node:} Depending on the determined intent, the process calls upon \texttt{translateToEN}, \texttt{translateToFrench}, or \texttt{translateToJP} to perform the translation.
    \item \textbf{End Node (end):} The conclusion of the process, where the translation result is outputted.
\end{itemize}

\begin{figure}[H]
\centering
\includegraphics[width=0.8\textwidth]{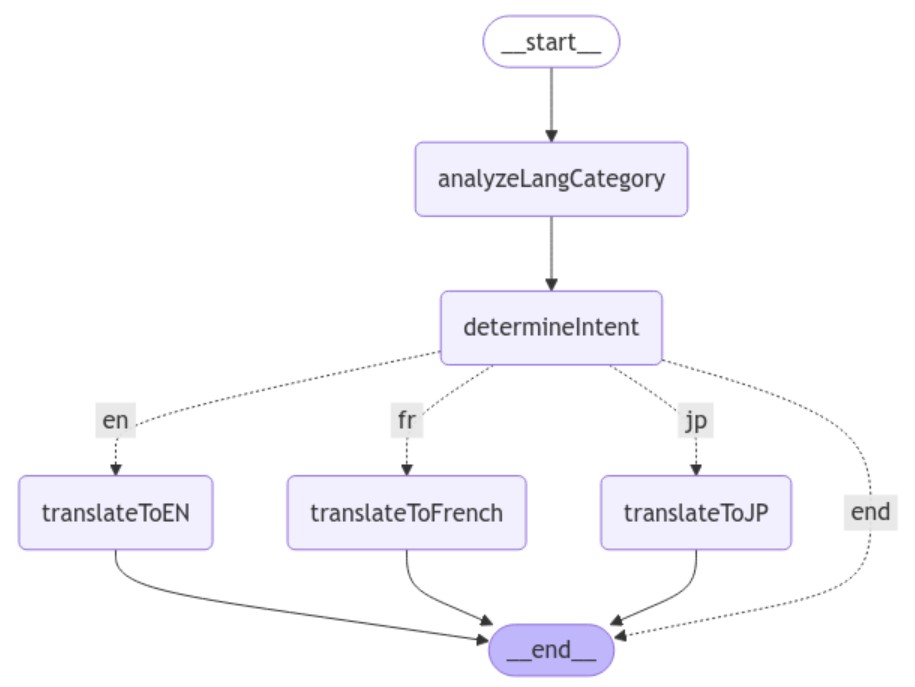}
\caption{Translation Agent Data Flowchart.} \label{fig7}
\end{figure} 
As shown in Figure 8, this study implemented a machine translation agent using the LangGraph framework and conducted translation experiments from English to French and from French to English. Figure (a) illustrates the translation experiment from English to French, while Figure (b) shows the translation experiment from French to English. The experimental results indicate that for translation tasks between English and French, the agent based on the LangGraph framework produced translation results that align with expectations, demonstrating the agent's ability to accurately comprehend the source language text and generate equivalent text in the target language. The experiments not only verified the agent's translation capabilities but also highlighted its potential for practical applications.

 \begin{figure}[H]
\centering
\includegraphics[width=1.0\textwidth]{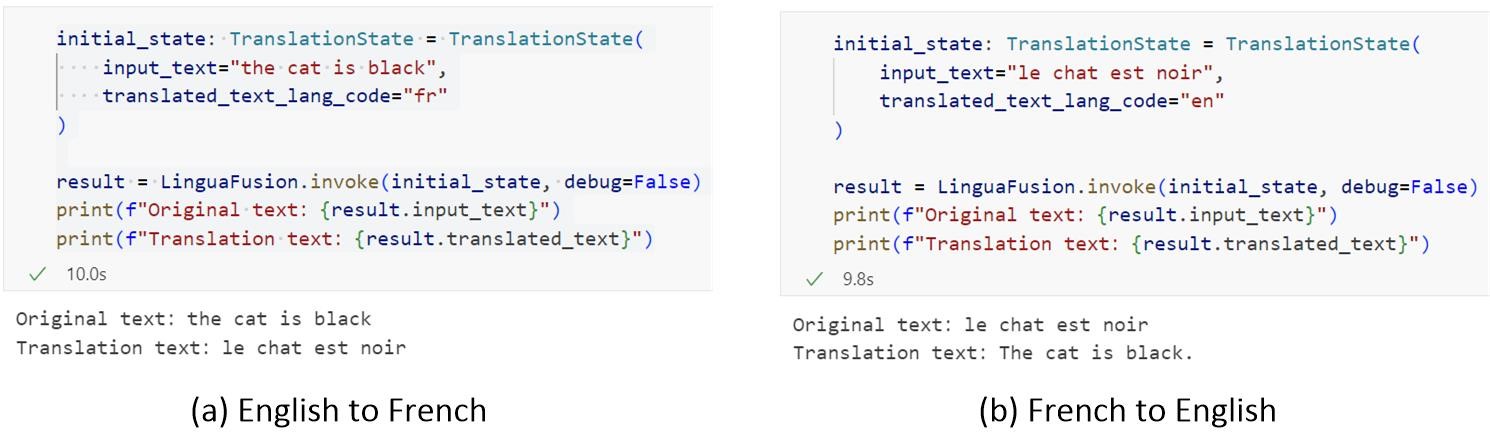}
\caption{English-French Translation Example.} \label{fig8}
\end{figure}

\subsection{Discussion}
Although this study has achieved positive results, there are still several areas that require further exploration and improvement:

1. Expanded Context Handling: Demonstrate LangGraph's ability to retain long-term context across paragraphs or entire documents rather than isolated sentences.

2. Dynamic Agent Adaptation: Showcase how agents can dynamically adapt to different domains or contexts using LangGraph, e.g., switching from casual conversational translation to technical or formal translation styles.

3. Interactive Features: Implement interactive feedback loops where users can provide corrections, and agents learn and adapt over time.

4. Human in the loop:Delve into the application of Human-in-the-loop within the LangGraph framework, setting breakpoints during the graph execution process to permit human intervention and adjustment. Engage in state editing and human feedback, modify the state of the graph during execution and adjust the workflow based on human input. Utilize the time travel feature, review the execution state of the graph by retracing its steps to facilitate debugging and optimization. This sets a course for future research in this domain.

\section{Conclusion}
This study underscores the significant potential of Agent AI and LangGraph in advancing machine translation technology. By leveraging the modularity and task-specific capabilities of agents alongside the robust workflow management provided by LangGraph, the proposed system achieves a high degree of flexibility, scalability, and accuracy in multilingual translation tasks. The integration of large language models (LLMs) like GPT-4o further enhances the semantic understanding and contextual relevance of translations.

LangGraph's graph-based framework plays a pivotal role in orchestrating these agents, enabling seamless state management, dynamic task allocation, and efficient multi-agent collaboration. This modular and automated design simplifies the addition of new languages and workflows, ensuring adaptability to evolving translation needs. Additionally, the system demonstrates the feasibility of maintaining contextual coherence and accuracy, essential for practical applications in diverse domains.

While the study achieved promising results, limitations in data availability and model complexity leave room for future exploration. Expanding the dataset, incorporating additional languages, and introducing features like long-term context retention and human-in-the-loop feedback will further refine the system’s capabilities.

The application of Agent AI with LangGraph highlights a new paradigm for machine translation, combining the strengths of advanced LLMs with modular, scalable frameworks. This approach not only drives improvements in translation quality but also paves the way for broader adoption of intelligent, adaptable language-processing solutions in diverse fields such as e-commerce, education, and international communication.

% \begin{thebibliography}{8}
 
% \end{thebibliography},

% \bibliographystyle{unsrt} 
 
% \bibliography{pytorch} 

\end{document}